%% file: main.tex
\documentclass[lettersize,journal]{IEEEtran}
\usepackage{amsmath,amsfonts}
\usepackage{algorithm}
\usepackage{array}
\usepackage[caption=false,font=normalsize,labelfont=sf,textfont=sf]{subfig}
\usepackage{textcomp}
\usepackage{stfloats}
\usepackage{url}
\usepackage{verbatim}
\usepackage{graphicx}
\usepackage{algpseudocode}

\usepackage[sorting=none, maxcitenames=1, maxbibnames=99, style=numeric-comp]{biblatex}
\AtBeginBibliography{\small}
\let\oldReturn\Return
\renewcommand{\Return}{\State\oldReturn}

\begin{document}

\title{I Can Hear You Coming: RF Sensing for \\ Uncooperative Satellite Evasion}

\author{Cameron Mehlman, Gregory Falco
\thanks{Dept. of Mechanical and Aerospace Engineering}%
\thanks{Cornell University, Ithaca, NY 14850, USA}}



\maketitle

\input{txt/abstract}

\section{Introduction}
\input{txt/intro}

\section{Related Works}
\label{sec:related}
\input{txt/related}

\section{Preliminaries and Problem Definition}
\label{sec:prelim}
\input{txt/prelim}

\section{Simulation}
\input{txt/sim}

\section{Method}
\label{sec:method}
\input{txt/method}

\section{Baselines}
\label{sec:baselines}
\input{txt/baselines}

\section{Experiments and Results}
\input{txt/experiments}

\input{txt/results}

\section{Discussion}
\input{txt/discussion}

\section{Conclusion}
\input{txt/conclusion}

\section*{Acknowledgments}
The authors world like to thank Adhyan Prasad for his support in initial simulation development.

\input{txt/appendix}

%

\printbibliography
\section{Biography Section}



\vspace{-33pt}
\begin{IEEEbiographynophoto}{Cameron Mehlman}
received a B.S. in aeronautical engineering from Rensselaer Polytechnic Institute (RPI) in 2021, and an M.S. in Mechanical Engineering specializing in robotics and controls from Columbia University in 2023. He is currently working towards a Ph.D. in mechanical engineering for Cornell University at the Aerospace ADVERSARY Lab.

His interests include applied Reinforcement Learning for planning of high degree of freedom mobile robotics.
\end{IEEEbiographynophoto}

\vspace{-33pt}

\begin{IEEEbiographynophoto}{Gregory Falco}
is an Assistant Professor at the Sibley School of Mechanical and Aerospace Engineering and the Systems Engineering Program at Cornell University. He received his PhD from MIT, where NASA’s Jet Propulsion Laboratory funded his doctoral research in Cybersecurity at MIT’s Computer Science and Artificial Intelligence Laboratory (CSAIL). Prior to joining Cornell University, he was an Assistant Professor at Johns Hopkins University’s Institute for Assured Autonomy and completed postdoctoral research at Stanford University’s Freeman Spogli Institute and MIT CSAIL.
\end{IEEEbiographynophoto}

\vfill

\end{document}

%% file: txt/abstract.tex
\begin{abstract}

This work presents a novel method for leveraging intercepted Radio Frequency (RF) signals to inform a constrained Reinforcement Learning (RL) policy for robust control of a satellite operating in contested environments. Uncooperative satellite engagements with nation-state actors prompts the need for enhanced maneuverability and agility on-orbit.  However, robust, autonomous and rapid adversary avoidance capabilities for the space environment is seldom studied. Further, the capability constrained nature of many space vehicles does not afford robust space situational awareness capabilities that can be used for well informed maneuvering. We present a ``Cat \& Mouse" system for training optimal adversary avoidance algorithms using RL. We propose the novel approach of utilizing intercepted radio frequency communication and dynamic spacecraft state as multi-modal input that could inform paths for a mouse to outmaneuver the cat satellite. Given the current ubiquitous use of RF communications, our proposed system can be applicable to a diverse array of satellites. In addition to providing a comprehensive framework for training and implementing a constrained RL policy capable of providing control for robust adversary avoidance, we also explore several optimization based methods for adversarial avoidance. These methods were then tested on real-world data obtained from the Space Surveillance Network (SSN) to analyze the benefits and limitations of different avoidance methods.

\end{abstract}

\begin{IEEEkeywords}
On-Orbit Simulation, Multi-Agent, Reinforcement Learning, constrained Reinforcement Learning, RF, Communications, Localization, Space, Adversary
\end{IEEEkeywords}

%% file: txt/intro.tex
In March of 2025, Chinese satellites exhibited dog-fighting capabilities \cite{McCarthy_2025}, following years of both Russian \cite{Hitchens_2023} and Chinese \cite{Jones_2023a} satellites approaching dangerously close to US satellites in geosynchronous orbit. Such uncooperative activity prompts the need for satellite agility and maneuverability which can be facilitated through edge-based autonomy. To achieve this, appropriate sensing would be required to properly characterize the contested environment. Not all satellites have precise space domain awareness (SDA) sensing suites onboard, despite having powerful buses and flight controllers that can facilitate autonomous operations. We propose leveraging an uncooperative space vehicle's communication systems as a means to evaluate safe flight control policies to carefully navigate contested domains in situations where support from the ground is not feasible.

\begin{figure}[htbp]
\centering
\framebox{\parbox{3.15in}{
\includegraphics[width=0.44\textwidth]{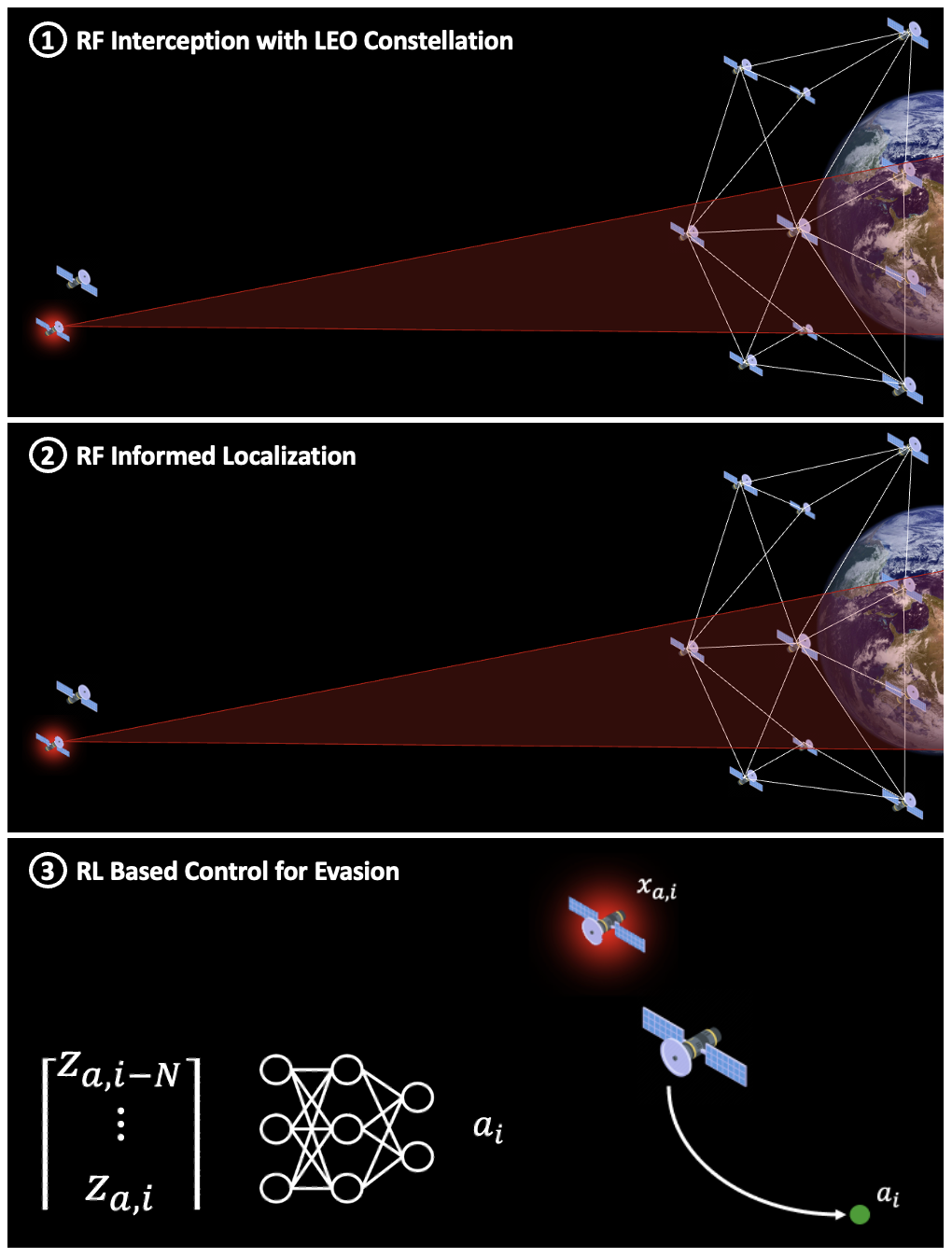}
        }}
\caption{Our proposed sensing and control architecture for a Cat and Mouse scenario.}
\label{fig:intro}
\end{figure}

Contested environment scenarios when engaging with an intelligent adversary in space is seldom discussed in the open literature. In part, this is a function of the historically poor maneuverability and sensing in space which is necessary for intelligent evasion. Recent investments in the `new space' industry have made such maneuverability more commonplace, thereby calling for novel intelligence capabilities to provide autonomous decision insight for these space vehicles. We propose a feedback architecture for uncooperative maneuvering that involves coupling an array of communications transceivers and a centralized reinforcement learning model to train a spacecraft to tactically maneuver around a non-cooperative satellite. We call our proposed scenario a "Cat \& Mouse" system which entails a ``mouse'' spacecraft that attempts to evade a second ``cat'' satellite's sensing capabilities while the ``cat'' pursues the ``mouse''. This generalized scenario is informed by recent events\cite{Hitchens_2023,Jones_2023a,McCarthy_2025}. The intention is to use the cat's communications capabilities sensed by collaborative assets in lower altitude orbits to estimate the cat's location and train a complex behavioral policy for contested space environments using a Reinforcement Learning (RL) approach as illustrated in Figure \ref{fig:intro}. 


Our motivation for utilizing RF sensors is to develop a system that engages a ubiquitous sensing modality across all satellites. This is in contrast to other traditional space situational awareness sensors such as optical or laser-ranging hardware which is only found on a subset of satellites. Moreover, the practice of locating unknown RF sources is well-studied, and several methods can be engaged to equip a nominally designed satellite with necessary data to autonomously maneuver in contested environments. Using these insights, we can more specifically define our cat \& mouse scenario. We assume that the cat and mouse are sufficiently far apart such that the mouse cannot sense the cat with on-board sensors. Additionally, we assume that the mouse spacecraft can be aided by receiving additional information from satellites in lower orbit intercepting the cat's RF communication signals. The mouse must then use this information to compute optimal spacecraft control.

In simulated environments, RL has proven useful to train generalized and adaptive policies for a range of different planning tasks. However, the behavioral policy learned is limited by the quality of the data it was trained on i.e. the quality of the simulation. Thus, to design a practical cat and mouse system as described, we first propose a simulation architecture that accounts for the orbital dynamics of a spacecraft as well as limitations of RF localization in order to accommodate a variety of scenarios and applicable spacecraft. The RL model is trained assuming the mouse satellite possesses no advanced adversary detection or sensing capabilities onboard and that the mouse must exclusively rely on RF to learn defensive behaviors. The final policy is then tested in a constrained manor on real-world telemetry data, and compared against standard optimization algorithms.

We propose a unique system architecture for adversary evasion that relies on sensing RF communications to achieve complex orbital maneuvers. The key contributions of this paper are:

\begin{itemize}
\item Define the framework for simulating intercepted radio frequencies from an adversarial spacecraft, as well as the key variables for location estimation and how to compute them in a dynamic space environment
\item Define the behavior of an adversarial satellite, and design a realistic simulation framework for training an evasive policy for this behavior
\item Design a Constrained Reinforcement Learning policy which incorporates both the dynamic spacecraft state as well as the processed RF observations in order to provide optimal evasive control for the mouse spacecraft
\item Analyze the performance of the constrained RL policy against standard collision avoidance methods currently used on satellites on real-world data
\end{itemize}

%% file: txt/related.tex
In recent years there has been growing attention focused on contested space, and satellite behavior when confronted with adversarial spacecraft.
\cite{prince2018,doi:10.2514/1.A35800,yang2024,}. However, these works often use optimization and heuristic based solutions which are only applicable within the narrow range of their target scenario, and lack the ability to develop display behaviors. Furthermore, autonomous capabilities in space are severely bottelnecked by our current space infrastructure's Space Sensing Awareness (SDA) capabilities, which cannot necessarily provide exact position or orbit of the adversary in real time \cite{8569343}. If autonomous satellite capabilities are to improve, new methods for sensing and detecting non-cooperative satellites must be explored.

Currently, the only widely available source of satellite orbital data is provided by the US Space Surveillance Network (SSN) \cite{nasa120Identification}. The SSN tracks approximately 45,000 satellites and pieces of debris using largely ground based radar. The SSN updates the orbit of each object in increments of one to several times a day while sensitive space assets' data are redacted. 
Other groups have shown the viability of using optics \cite{https://doi.org/10.1029/JB088iB01p00669,RIEL2019121} and laser ranging \cite{Wilkinson2019} to track objects in orbit, however, both methods require an extensive ground infrastructure as well as prior knowledge of the object of interest. While all of the methods discussed have provided sufficient data for current satellite needs, none show scalable potential for providing real-time updates of non-cooperative satellite behavior for collision avoidance algorithms run on the edge. Our work proposes to leverage intercepted Radio Frequency data from a non-cooperative satellite to estimate its location in real time as observation to inform a control algorithm.

Radio Frequency data has long been used to calculate the position and trajectory of cooperative satellites \cite{Guo2010,https://doi.org/10.1002/navi.281,7341285,7341285}, and is often a key component of any satellite Telemetry, Tracking, \& Command (TT\&C) system. However, the concept of locating non-cooperative satellites using RF from space is a fairly unexplored topic. Al-Hourani et. al. proposes a method to estimate the distance between two satellites based off of the Doppler effect \cite{article24}, but the method does not attempt to estimate the actual location of the unknown satellite. Others have proposed leveraging RF receivers in Low Earth Orbit (LEO) for geolocation \cite{10356280,7134744} although assume a static emitter on Earth's surface. This being said, there are a number of works analyzing the different types of methods than can be used on Earth for location estimation of an unknown adversary's position using RF. Some of the most popular methods utilize features such as Received Signal Strength (RSS) \cite{301830,8974236,8240634,6850028}, and a combination of the Time Difference on Arrival (TDOA) and Frequency Difference on Arrival (FDOA) \cite{1323254,7880621,7582404,s23146254,ZHOU2021109758}. Although these methods do not address the complexities of the space domain, they highlight the key variables that are necessary in order to infer essential location information of an unknown emitter. 

In addition to potential sensing modalities for space domain awareness (SDA), it is important to recognize the current standards for satellite planning and their limitations. The majority of proposed works frame satellite maneuvering as a complex optimization problems that assumes accurate telemetry data of the satellite as well as its environment \cite{8187728}. Luo et. al. proposes a novel adaptive differential evolutionary algorithm for optimizing earth-observation satellite maneuvers \cite{rs14091966}. Porcelli et. al. provide a method for detecting and estimating satellite maneuvers using radar observations assuming an observation accuracy of 5 meters \cite{PORCELLI2022274}. Hu et al. proposes a complex mathematical method for path planning in space environments with close proximity to obstacles, but again require full knowledge of the satellites surrounding state \cite{Huxie}. While all of the discussed proposed methods satisfy the requirements for specific space missions, their computational complexity and inability to handle noisy observations highlights their inability to perform on the edge and with RF informed observations which can be significantly noisier than radar, laser, or optics.


Reinforcement Learning (RL) has show significant promise towards training complex control polices in the last few decades \cite{khandate2023dexterous,levine2016endtoend,mnih2013atari}, and some have begun to implement Reinforcement Learning for the space environment in order to train optimal policies for communications planning in contested environments \cite{stimpson2015,10015325}, attitude control \cite{osti_10156483,s18124331}, and maneuvering \cite{TIPALDI20221}. In order to address in deployment of RL models, Constrained Reinforcement Learning (CRL) -- sometimes referred to as shielded RL -- has become popular for ensuring the RL agent only takes safe actions during implementation \cite{10423180,banerjee2024dynamicmodelpredictiveshielding,10.1145/3568162.3576983,10.5555/3504035.3504361,achiam2017constrainedpolicyoptimization}, and shows significant promise for higher risk environments such as space. However, there has been limited work exploring the potential for RL to produce policies that can guide a spacecrafts physical behavior when performing in contested environments. A number of works have investigated RL control policies for individual or collaborative inspection of potentially non-cooperative or non-operational spacecraft \cite{10.1007/978-3-031-51928-4_76,vanwijk2023deepreinforcementlearningautonomous,doi:10.2514/1.I011391}. Others have explored the potential for RL based collision avoidance algorithms \cite{dunlap2024deepreinforcementlearningscalable,10660520,bourriez2023spacecraftautonomousdecisionplanningcollision}, although none of the mentioned works attempt to design policies capable of performing in contested environments which require higher behavioral capabilities and provide shorter windows to act. This leaves a large gap in work that attempts to leverage current space sensing capabilities in order to train reactive, robust policies for adversarial evasion.

Our work proposes a new model for edge based satellite autonomy which addresses shortcomings both in SDA and modern satellite maneuvering algorithms by exploring new sensing modalities and the ability of RL policies to provide robust and optimal control. This method leverages RF transmissions intercepted in LEO to estimate the location of a potential adversary. We then show that this observational data can be utilized by a Reinforcement Learning policy to provide highly reactive and optimized control, keeping the potential adversary at a desired distance, while minimizing fuel cost and deviation from the initial orbit.



%% file: txt/prelim.tex
We will first describe the preliminary information required to frame the RF informed evasive planning problem. This first entails a formal definition of the multiple frames necessary to describe the behavior and motion of the multiple types of spacecraft involved in the problem statement. Then, we provide a brief overview of Markov Decision Processes (MDPs) and the framing of the proposed problem statement as such.

\subsection{Reference Frames}

\begin{figure}[htbp]
\centering
\framebox{\parbox{3.4in}{
\includegraphics[width=0.47\textwidth]{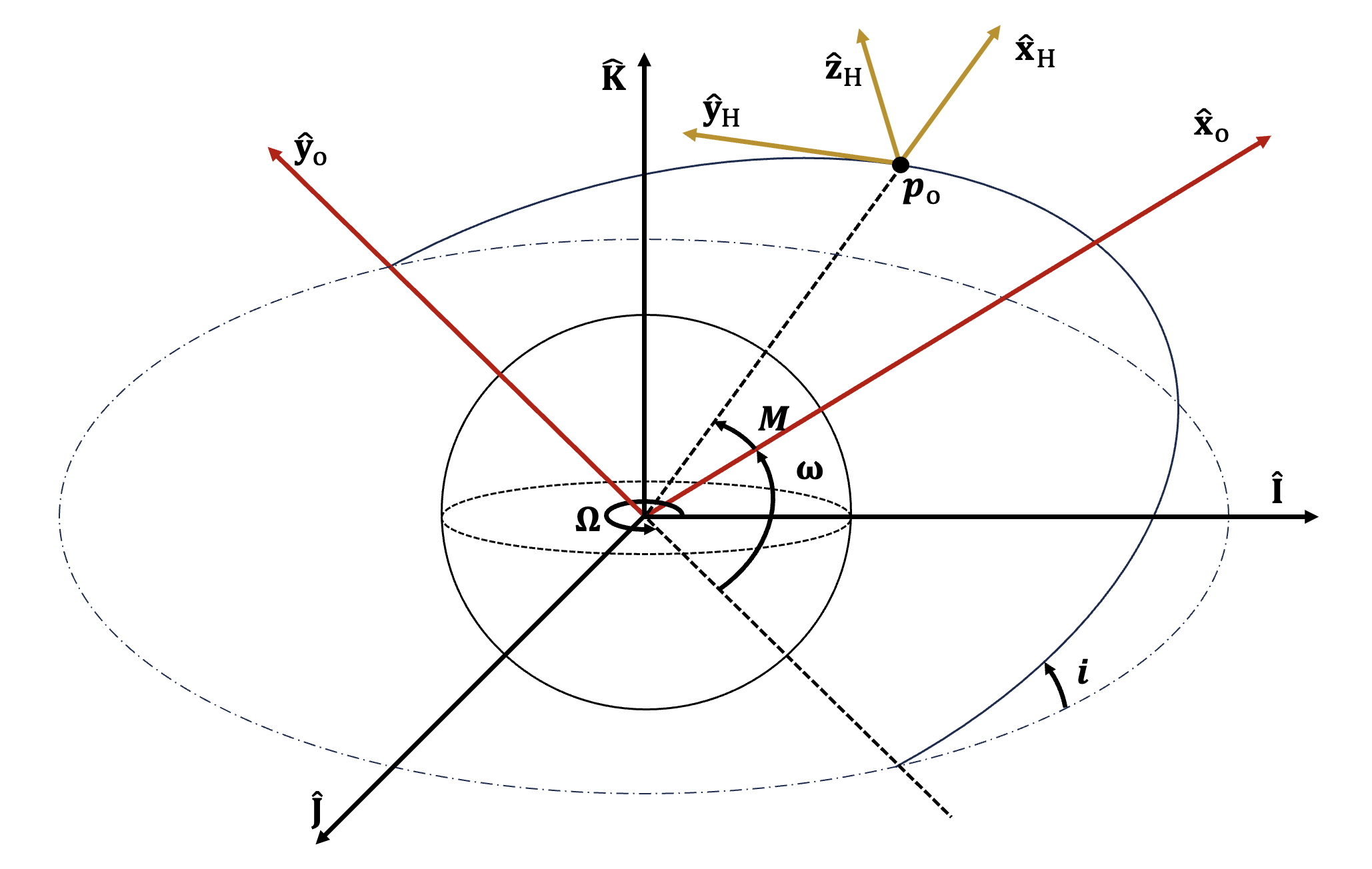}
        }}
\caption{The ECEF (black), orbital (red), and Hill (yellow) reference frames, as well as orbital angles ($\Omega, i, \omega$) and mean anomaly ($M$) where $\boldsymbol{p}_o$ is the point on the orbit that serves as the origin to the Hill frame.}
\label{fig:frames}
\end{figure}

In order to model the problem, the dynamics of the system are described in several reference frames: first, the inertial frame or Earth-Centered, Earth-Fixed (ECEF) notated by $\hat{\text{\textbf{I}}},\hat{\text{\textbf{J}}},\hat{\text{\textbf{K}}}$, second, the orbital frame notated by $\hat{\text{\textbf{x}}}_o,\hat{\text{\textbf{y}}}_o,\hat{\text{\textbf{z}}}_o$, and finally, the localized frame or Hill frame notated by $\hat{\text{\textbf{x}}}_H,\hat{\text{\textbf{y}}}_H,\hat{\text{\textbf{z}}}_H$. An image depicting the three frames and their relations can be seen in Figure \ref{fig:frames}. In simulation, coordinates are generally kept in the ECEF and Hill frame, translating between the two frames can be done by:

\begin{align}
\boldsymbol{p}^H = R_{OH}R_{EO} (\boldsymbol{p}^{E} - \boldsymbol{p}^H_O)
\label{eq:translate}
\end{align}

where $\boldsymbol{p}^{H}$ is the translated point in the Hill frame, $\boldsymbol{p}^{E}$ is a point in the ECEF Frame, $\boldsymbol{p}^H_O$ is the origin of the Hill Frame, $R_{EO}$ is a rotation matrix that translates from ECEF to orbital frame:

\begin{align}
R_{EO} = 
\begin{bmatrix}
    \boldsymbol{c}_\Omega\boldsymbol{c}_\omega - \boldsymbol{s}_\Omega\boldsymbol{c}_i\boldsymbol{s}_\omega &
    \boldsymbol{s}_\Omega\boldsymbol{c}_\omega + \boldsymbol{c}_\Omega\boldsymbol{c}_i\boldsymbol{s}_\omega &
    \boldsymbol{s}_i\boldsymbol{s}_\omega \\ 
    -\boldsymbol{c}_\Omega\boldsymbol{s}_\omega - \boldsymbol{s}_\Omega\boldsymbol{c}_i\boldsymbol{c}_\omega &
    -\boldsymbol{s}_\Omega\boldsymbol{s}_\omega + \boldsymbol{c}_\Omega\boldsymbol{c}_i\boldsymbol{c}_\omega &
    \boldsymbol{s}_i\boldsymbol{c}_\omega \\
    \boldsymbol{s}_\Omega\boldsymbol{s}_i &
    -\boldsymbol{c}_\Omega\boldsymbol{s}_i &
    \boldsymbol{c}_i
\end{bmatrix}
\label{eq:dcm eo}
\end{align}

and finally $R_{OH}$ is a rotation matrix that translates from the orbital frame to Hill Frame:

\begin{align}
R_{OH} = 
\begin{bmatrix}
    \boldsymbol{c}_M & -\boldsymbol{s}_M & 0 \\ 
    \boldsymbol{s}_M & \boldsymbol{c}_M & 0 \\
    0 & 0 & 1 \\
\end{bmatrix}
\label{eq:dcm oh}
\end{align}

where $i, \omega, \Omega$ are the orbital angles inclination, argument of periapsis, and right ascension of ascending node (RAAN) respectively. And, $M$ is the mean anomaly of the orbit. Additionally, it should be noted that in Equations \ref{eq:dcm eo} and \ref{eq:dcm oh} the notation $\boldsymbol{c}_\Omega$ represents $cos(\Omega)$. Each spacecraft can be tracked in all three frames at once, however, spacecraft that are assumed not to have thrust control such as LEO constellation assets are only described in the ECEF and orbital frame.

\subsection{MDP Problem Definition}

In order to solve for the sequential decision making process required for reactive evasive maneuvering we model the system as an infinite horizon MDP which provides a framework for describing stochastic environments with nondeterministic actions by state-action pairs \cite{bellman1,bellman2}. An infinite horizon MDP can be formally defined by $\mathcal{M}=(\mathcal{S},\mathcal{A},T,s_0,R,\gamma)$, where $\mathcal{S}$ is the state space of the environment, $\mathcal{A}$ is the action space, $T:\mathcal{S}\times\mathcal{A}\times\mathcal{S}$ is the transition function, $s_0:\mathcal{S}$ is the initial state, $R:\mathcal{S}\times\mathcal{A}\times\mathcal{S}$ is the reward function, and $\gamma\in [0,1)$ is a discount factor. At each discrete timestep, an agent receives the current state of the environment $s$, and follows a policy $\pi_\phi$ to chose an action $a$. The agent then receives a reward $r$ and is informed of the resulting state $s'$. An optimal policy for the MDP $\pi_\phi^*:\mathcal{S}\times\mathcal{A}$ chooses actions that maximize the expected discounted utility:

\begin{align}
\pi_\phi^* = \underset{a\in\mathcal{A}}{\operatorname{argmax}} Q^*(s,a)
\label{eq:qlearning}
\end{align}

where $Q^*(s,a)$ is the expected utility of taking action $a$ at state $s$. Critical to learning the optimal policy is mapping the transition function $T(s,a,s^\prime) \forall s\in\mathcal{S},a\in\mathcal{A},s^\prime\in\mathcal{S}$, which provides the probability of transitioning to state $s^\prime$ given you take action $a$ from state $s$. The dynamics of both the cat and mouse spacecraft are modeled as linear and time discrete:

\begin{align}
\begin{bmatrix}
\boldsymbol{x}^H_{i+1} \\
\dot{\boldsymbol{x}}^H_{i+1}
\end{bmatrix}
 = 
 A
\begin{bmatrix}
\boldsymbol{x}^H_{i} \\
\dot{\boldsymbol{x}}^H_{i}
\end{bmatrix}
+ B 
\begin{bmatrix}
\boldsymbol{0} \\
\boldsymbol{u}_i
\end{bmatrix}
\end{align}

where $\boldsymbol{x} \in \mathbb{R}^3$ represents the position of the spacecraft in cartesian coordinates, $\dot{\boldsymbol{x}}\in \mathbb{R}^3$ represents the translational velocities, and $\boldsymbol{u}_i\in \mathbb{R}^3$ is the control computed by $f(\boldsymbol{x}_i,\dot{\boldsymbol{x}}_i,\boldsymbol{a}_i):\mathbb{R}^9\rightarrow\in \mathbb{R}^3$ which converts a desired change in position $\boldsymbol{a}_i$ to thrust control $\boldsymbol{u}_i$. Therefore, the full state at any point can be described by a timeseries of recent cat position estimates, dynamic state of the mouse spacecraft, and current desired location:

\begin{align}
\boldsymbol{s}_i = [\boldsymbol{x}^H_{m,i},\dot{\boldsymbol{x}}^H_{m,i},\boldsymbol{a}_{i}+\boldsymbol{x}^H_{m,i},{\boldsymbol{x}^H_{a,i-N}},...,{\boldsymbol{x}^H_{a,i}}]
\end{align}

where the subscript $a$ represents the adversary or cat, and the subscript $m$ indicates the mouse. Although the derivative of the cat's position is not explicitly provided, it is implicitly included in the time series data. The action $\boldsymbol{a}_i$ is then represented by the desired change in cartestian coordinates in the Hill frame:

\begin{align}
\boldsymbol{a} = [\Delta x^H, \Delta y^H, \Delta z^H]
\label{eq:action}    
\end{align}

Additionally, the received state of the cat is a noisy observation $z:\mathcal{S}$ which is an estimate of the true location with gaussian noise:

\begin{align}
{{\boldsymbol{z}_{a,i}}^H} = \mathcal{N}({\boldsymbol{x}^H_{a,i}}, \Sigma_t)
\end{align}

Thus, the problem is a partially-observable markov decision process (POMDP). POMDPs are well studied and can be considerably more complex to learn an optimal policy for \cite{NEURIPS2021_d5ff1353,DBLP:journals/corr/abs-2110-05038,DBLP:journals/corr/abs-2110-05038}. However, because the observed state $\boldsymbol{z}_{a,i}$ closely represents the true state, the system is assumed to behave as an MDP with gaussian noise in the observation -- similar to many domain randomization methods \cite{5979644,DBLP:journals/corr/abs-2110-03239,8202133}.



%% file: txt/sim.tex
In order to train a successful policy for evasion a simulation environment that accurately emulates the contested space environment as well as the sensor data collected is essential. In this section we define the framework for our simulation which models the orbital dynamics of each spacecraft, sensor constellation configurations, and finally RF localization.

\subsection{Orbital Dynamics Modeling}

The dynamic modeling of our simulation is composed of two layers: first the orbital dynamics of all satellites can be modeled using Kepler's Laws of Planetary Motion, and second the localized dynamics of the cat and mouse spacecraft are modeled using the Clohessy Wiltshire Equations. All orbits are assumed to be near circular, and therefore each one can be defined by five parameters: inclination ($i$), argument of periapsis ($\omega$), right ascension of ascending node (RAAN) ($\Omega$), semi-major axis ($a$), and mean anomaly (M).

Tracking the orbits of each satellite is necessary to accurately simulate the behavior of the intercepted signal frequency, which is dependent on the satellites' relative speed and position to the RF receivers which are configured in a Low Earth Orbit (LEO) constellation. By assuming a near-circular orbit, we are able to employ a simplified orbital model:

\begin{align}
M = nt
\label{eq:true anomaly}
\end{align}

where n is the mean motion:

\begin{align}
n = \sqrt{\frac{\mu}{a^3}}
\label{eq:mean motion}
\end{align}

and $\mu$ is Earth's gravitational parameter. Thus, the position of the satellite on its orbit can be computed in the orbital plane by:

\begin{align}
x = acos(M)\hat{\text{\textbf{x}}}_o
\label{eq:orbit posx}
\end{align}

\begin{align}
y = acos(M)\hat{\text{\textbf{y}}}_o
\label{eq:orbit posy}
\end{align}

and velocity can be computed as:

\begin{align}
v_x = \sqrt{\frac{\mu}{a}}sin(M)\hat{\text{\textbf{x}}}_o
\label{eq:orbit velx}
\end{align}

\begin{align}
v_y = \sqrt{\frac{\mu}{a}}cos(M)\hat{\text{\textbf{y}}}_o
\label{eq:orbit vely}
\end{align}

Equations \ref{eq:true anomaly}-\ref{eq:orbit vely} allow us to solve for the position and velocity of the a spacecraft's orbit at any point in time during the simulation.

In a localized frame the state of our cat and mouse spacecraft can be described using the Clohessy Wiltshire (CW) Equations. These equations model the behavior of a spacecraft in relation to a single point moving on a near circular orbit. We can track the location of this point using Equations \ref{eq:true anomaly}-\ref{eq:orbit vely}. The equations of motion for the two spacecrafts position relative to this point is then:

\begin{align}
\ddot{x} = \left( 3n^2x + 2n\dot{y} + \frac{T_x}{m} \right) \hat{\text{\textbf{x}}}_H
\label{eq:cwhx}
\end{align}
\begin{align}
\ddot{y} = \left( -2n\dot{x} + \frac{T_y}{m} \right) \hat{\text{\textbf{y}}}_H
\label{eq:cwhy}
\end{align}
\begin{align}
\ddot{z} = \left( -n^2z + \frac{T_z}{m} \right) \hat{\text{\textbf{z}}}_H
\label{eq:cwhz}
\end{align}

where $T$ is the thrust applied at each timestep in the respective axis. The equations describe translational motion in the Hill frame which has its x-axis tangential to the orbit of the origin, and y-axis normal to its orbit. Additionally, it is worth noting that the CW equations are a linearization of Newton's law for gravitational motion, and thus lose accuracy as the distance between the spacecraft and origin increase. Due to this, the CW equations are only used to model the dynamics of each spacecraft when within several tens of kilometers of the origin. 

\subsection{RF Modeling and Constellation Configuration}

\begin{figure}[htbp]
\centering
\framebox{\parbox{3.4in}{
\includegraphics[width=0.49\textwidth]{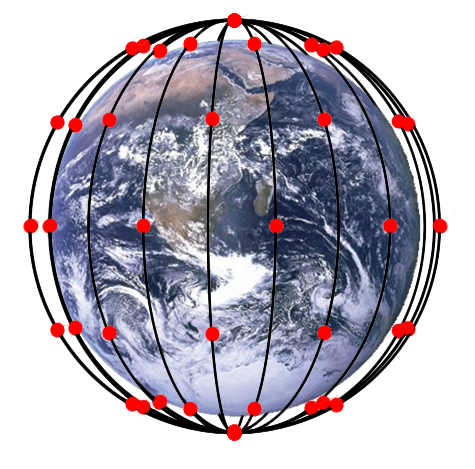}
        }}
\caption{An image depicting the constellation configuration used, where the red dots resemble a single satellite in the constellation.}
\label{fig:constellation}
\end{figure}

As established we assume there are two main agents: the cat and mouse which are both modeled in the localized Hill frame. However, the mouse is able to gain information received from several support satellites in a lower altitude constellation which are intercepting the cat's RF transmissions in order to estimate the location of the source. Thus, our simulation involves $M$ sensors in a 3D environment with a single RF source. For this experiment, all sensors are assumed to exist within the same LEO constellation. We simulate a standard Walker Constellation \cite{CORNARA2001681} in which all satellites are in circular polar orbits of the same altitude evenly distributed across $P$ orbital planes of varying RAAN. An image of this configuration can be seen in Figure \ref{fig:constellation}

Additionally, we assume that the RF transmission is directional. Thus, in order for a satellite to intercept the adversary's RF it must be within the broadcasted beam's width. In order to compute this, we model the beam as a conical shape defined by a unit vector $\hat{\boldsymbol{b}}$ which represents the center line of the cone, and angle $\theta$. Whether a satellite is within the beam width or not can than be calculated using:

\begin{align}
\phi = sin^{-1}\biggl(\frac{(\boldsymbol{x}^E_a  - \boldsymbol{x}^E_i)\times\hat{\boldsymbol{b}}}{||\boldsymbol{x}^E_a - \boldsymbol{x}^E_i||}\biggl)
\label{eq:beam}
\end{align}

where $\boldsymbol{x}^E_a$ is the location of the cat in the ECEF frame, and $\boldsymbol{x}^E_i$ is the location of a constellation satellite capable of sensing the RF transmission in the ECEF frame.

RF data can be analyzed through a number of different methods, in this work, we propose the utilization of a TDOA localization algorithm in order to estimate the position of the cat. In simulation, the TDOA can be computed by:

\begin{align}
\tau_i = \frac{\Delta d_i}{c} + \mathcal{N}(0,\,\sigma_d^{2})
\label{eq:tdoa}
\end{align} 

where $c$ is the speed of propagation, $\mathcal{N}(0,\,\sigma_d^{2})$ is some gaussian noise with mean 0 and variance of $\sigma_d^{2}$, and $\Delta d_i$ is the range difference:

\begin{align}
\Delta d_i = || \boldsymbol{x}^E_a - \boldsymbol{x}^E_i || - || \boldsymbol{x}^E_a - \boldsymbol{x}^E_1 ||
\label{eq:d}
\end{align}

In Equation \ref{eq:d} $\boldsymbol{x}^E_1$ is the location of the reference sensor.


\subsection{Cat Localization}

In order to effectively plan around an uncooperative satellite, it is essential that some form of localization is provided to the mouse spacecraft which is done in this work by implementing a TDOA localization algorithm. As discussed in Section \ref{sec:related}, a large number of different TDOA localization algorithms have become standard practice in a wide range of applications including cellular networks, law enforcement, and indoor positioning \cite{8580464}. Generally, the performance of a TDOA localization algorithm is measured by its ability to perform close to its theoretical Cramer-Rao Lower Bound, which was derived by K.C. Ho et al. \cite{599239}:

\begin{align}
\text{CRLB} = \text{\textbf{J}}_{\text{TDOA}}^{-1}
=
\left(\frac{\partial\boldsymbol{d}}{\partial\boldsymbol{u}}^T\text{\textbf{Q}}_f^{-1}\frac{\partial\boldsymbol{d}}{\partial\boldsymbol{u}}\right)^{-1}
\label{eq:crlb}
\end{align}

\begin{align}
\frac{\partial\boldsymbol{d}}{\partial\boldsymbol{u}} = 
\begin{bmatrix} 
    \frac{\boldsymbol{x^E}^T_2 - \boldsymbol{x^E}^T_a}{r_2} - \frac{\boldsymbol{x^E}^T_1 - \boldsymbol{x^E}^T_a}{r_1}\\
    \frac{\boldsymbol{x^E}^T_3 - \boldsymbol{x^E}^T_a}{r_3} - \frac{\boldsymbol{x^E}^T_1 - \boldsymbol{x^E}^T_a}{r_1}\\
    \vdots \\
    \frac{\boldsymbol{x^E}^T_M - \boldsymbol{x^E}^T_a}{r_M} - \frac{\boldsymbol{x^E}^T_1 - \boldsymbol{x^E}^T_a}{r_1}\\
    \end{bmatrix}
\label{eq:crlb2}
\end{align}

where $\text{\textbf{Q}}_f$ is the noise matrix:

\begin{align}
\text{\textbf{Q}}_f = 
\frac{1}{2}c^2\sigma_d^2 (\text{\textbf{I}}_{M-1} + \boldsymbol{1}_{M-1}) 
\label{eq:noise matrix}
\end{align}

and:

\begin{align}
r_i = || \boldsymbol{x}^E_a - \boldsymbol{x}^E_i ||
\label{eq:sensor range}
\end{align}

Multiple works cited in Section \ref{sec:related} have shown their methods to perform similar to the CRLB both in simulation and real world testing \cite{ZHOU2021109758,s23146254,7582404,10356280}. Thus, in order to reduce computational complexity of the simulation the CRLB is computed using Equations \ref{eq:crlb}-\ref{eq:sensor range} in order to provide an appropriate noise level for the cat's location estimate rather than run a TDOA localization algorithm using simulated TDOA data every timestep. The corresponding position estimate can be simulated by summing a sampled error using the CRLB and the true position of the cat:

\begin{align}
{\boldsymbol{z}^E_a} = \boldsymbol{x}^E_a + \mathcal{N}(0,\,\sqrt{diag(\text{CRLB})})
\label{eq:cat est}
\end{align}

\begin{figure}[htbp]
\centering
\framebox{\parbox{3.4in}{
\includegraphics[width=0.49\textwidth]{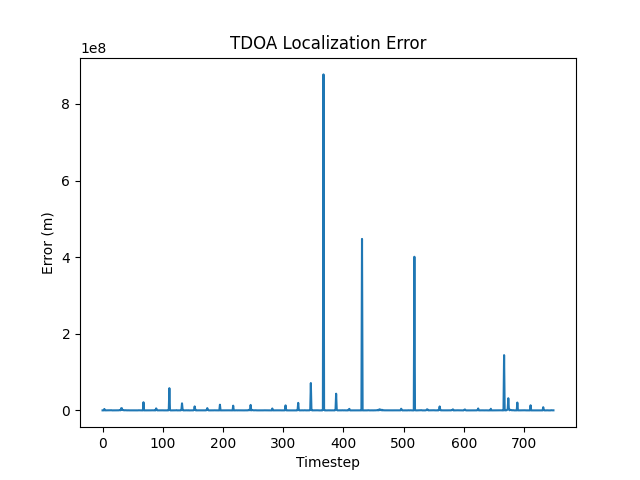}
        }}
\caption{The simulated error between the cat position estimated by a TDOA localization algorithm and its true position during a single run of the simulation.}
\label{fig:tdoa error}
\end{figure}

Due to the distance between the RF sensors and source as well as the fact that the sensors are constantly moving in relation to the source the size of the noise can vary from nominal to extreme. A graph depicting the error of the position estimates during a single episode can be seen in Figure \ref{fig:tdoa error}. For this reason, we find it necessary to incorporate a filter into our system in order to properly process the observation data which will be discussed in Section \ref{sec:method}.

%% file: txt/method.tex
In this section we will discuss our proposed control scheme of evasive maneuvering of the mouse. In order to properly evade an uncooperative actor while maintaining minimal deviation from the initial path an agent must be capable of perceiving its surroundings, processing observations, computing optimal actions, and finally executing said actions in a single feedback loop. Control of the spacecraft can be broken into two stages: the RL policy which receives an observation state of the environment and computes the a desired change in position, and an MPC controller (previously defined as $f(\boldsymbol{x}_i,\dot{\boldsymbol{x}}_i,\boldsymbol{a}_i)$) which converts positional commands to thrust control. In addition, we implement an Extended Kalman Filter during testing in order to improve accuracy of the observations received and overall performance of the constrained RL policy.


\subsection{Reinforcement Learning Training Scheme}

The RL policy is tasked with computing optimal positional control given some observation of the environment. The observation provided to the cat is a concatenation of the mouse's state in the local frame, the mouse's current action, and the estimated state of the cat in the local frame from the last $N$ timesteps:

\begin{align}
\boldsymbol{o}_i = [\boldsymbol{x}^H_{m,i},\dot{\boldsymbol{x}}^H_{m,i},\boldsymbol{a}_{i}+\boldsymbol{x}^H_{m,i},{\boldsymbol{z}^H_{a,i-N}},...,{\boldsymbol{z}^H_{a,i}}]
\end{align}



As described by Equation \ref{eq:action} $\boldsymbol{a}_i$, is the desired change in current location of the mouse. This is then used to compute the desired location of the mouse in the local coordinate frame, and an MPC is utilized to compute the necessary thrust control to be applied. 

In order to gauge the utility of an action during training, the policy is provided feedback in the form of a reward. During training, the reward function implemented prioritizes deviation from the initial orbit, fuel consumed, and proximity to the cat:

\begin{align}
r_i = 
\Biggl\{
\begin{array}{lr}
 \text{if } \Delta d_c > d_{tol} & 1 - c_1||\boldsymbol{x}^H_{m,i}|| - c_2f_i\\
 \text{else } & 0
\end{array}
\label{eq:rew}
\end{align}

where $c_1,c_2$ are constants, and $f_i$ is the fuel consumed since the previous timestep. The reward function is bounded to $[0,1]$, thus, encouraging the mouse to take actions that minimize deviation from the origin and fuel while staying a minimum distance of $d_{tol}$ from the cat. During all experiments and training, $d_{tol}$ is set to 20km.


As discussed in Section \ref{sec:prelim}, the noisy position estimates of the cat prevents the policy from knowing the true state of the environment at any given time. And, while this can be used to improve generalization of the policy during testing, it can also hinder the chance of learning an optimal policy. In order to mitigate this a curriculum learning strategy is applied to the localization noise during training, similar to many curriculum-based domain randomization techniques \cite{DBLP:journals/corr/abs-1910-07113,10606428}. In order to implement this Equation \ref{eq:cat est} is adapted to:

\begin{align}
{\boldsymbol{z}^E_a} = \boldsymbol{x}^E_a + \alpha\mathcal{N}(0,\,\sqrt{diag(\text{CRLB})})
\label{eq:noise growth}
\end{align}

where $\alpha$ is a constant that follows a piecewise growth scheduled. By implementing this, the RL policy is able to first converge upon an optimal policy with exact position observations of the cat before having to generalize to incorporate noise. 

\begin{figure}[htbp]
\centering
\framebox{\parbox{3.4in}{
\includegraphics[width=0.47\textwidth]{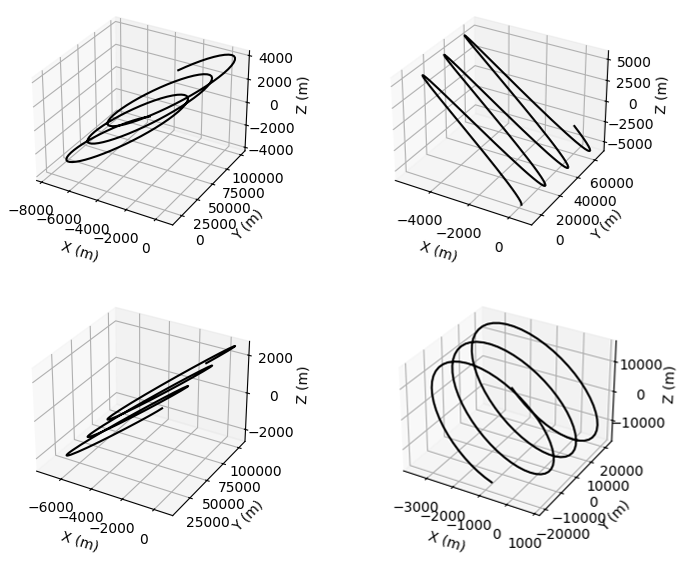}
        }}
\caption{The trajectories the cat spacecraft during several randomly generated scenarios in simulation.}
\label{fig:cat traj}
\end{figure}

Additionally, it is important during training that the adversary's actions represent a sufficiently diverse set of potential behaviors so that the trained policy is able to demonstrate robust and generalizable capabilities during testing. In order to ensure this the cat is modeled as a chaser spacecraft with no thrust control. The spacecraft then drifts in the vicinity of the mouse, requiring the mouse to adjust its position to maintain a minimum distance. Depending on the initial position and velocity of the chaser spacecraft in the Hill frame, its path can vary drastically. An image depicting several different paths from randomly generated episodes can be seen in Figure \ref{fig:cat traj}.

\subsection{Constrained Reinforcement Learning}

It is essential that during testing the actions provided by the RL policy always performs within acceptable bounds. Thus, we constrain the potential action set during testing by weighting the probability of taking specific actions. In order to do this, we have identified three potential action scenarios for the cat and mouse simulation: first, the cat is within a predetermined distance tolerance of the mouse and evasive maneuvers should be taken ($d_c < c_1$), second, the cat is outside of the first distance tolerance but not sufficiently far and thus best to wait to understand its behavior ($c_1 < d_c < c_2$), and third, the cat is sufficiently far from the mouse and the spacecraft should return to its initial orbit ($d_c > c_2$). This allows us to define the potential action space during testing:

\begin{align}
\boldsymbol{A}_i = 
\Biggl\{
\begin{array}{lr}
\sim \pi_\phi(\boldsymbol{a}_i\mid \boldsymbol{o}_i) \\
 \boldsymbol{0} \\
 -\boldsymbol{x}^H_{m,i}
\end{array}
\label{eq:test action space}
\end{align}

where $\sim \pi_\phi(\boldsymbol{a}_i\mid \boldsymbol{o}_i)$ is the action sampled from the RL policy, $\boldsymbol{0}$ would entail no change in the mouse's position, and $-\boldsymbol{x}^H_{m,i}$ is the necessary change in position to return to the origin. Each action represents the scenarios mentioned in their respective order. At each timestep during testing actions are sampled following discrete probability distribution which represents the probability of each of the three previously defined inequalities being true ($\pi_{s1},\pi_{s2},\pi_{s3}$).


Each probability is computed by taking the weighted sum of the probability of each inequality for the past $N$ estimates of the cat's position:

\begin{align}
{\pi_s}_k = \sum_{j=1}^N w_j P(b\mid{{\boldsymbol{z}_{a,i-N}^H}^+},\boldsymbol{x}_{m,i}^H)
\label{eq:summed prob}
\end{align}

where $b$ represents one of the three potential inequalities for each scenario, and $w$ comes from a set of normalized weights. The function $P(b\mid{{\boldsymbol{z}_{a,i-N}^H}^+},\boldsymbol{x}_{m,i}^H)$ is the cumulative density function (CDF) of the euclidean distance between the cat and mouse given an observed location. This can be computed by treating it as a noncentral chi-squared distribution with three dimensions:

\begin{equation}
\begin{aligned}[b]
& P(b\mid \boldsymbol{x}) = 1  \\
& - \left[ Q(\sqrt{\hat{c}}-\sqrt{\lambda}) + Q(\sqrt{\hat{c}}+\sqrt{\lambda}) + \sqrt{\frac{2}{\pi}} \frac{sinh(\sqrt{\lambda \hat{c}})}{\sqrt{\lambda}} e^{-\frac{\hat{c}+\lambda}{2}} \right]
\end{aligned}
\end{equation}

where $\boldsymbol{x}$ is the mean of the point of interest equal to ${{\boldsymbol{z}_{a,i-N}^H}^+}-\boldsymbol{x}_{m,i}^H$, $Q()$ is a gaussian Q function:

\begin{align}
Q(x) = \frac{1}{2}\left(\frac{2}{\sqrt{\pi}}\int_{x/\sqrt{2}}^\infty e^{-t^2}dt\right)
\label{eq:gaussianq}
\end{align}

$\hat{c}$ is the square of the distance tolerance normalized by the standard deviation $\sigma$ of the position estimate:

\begin{align}
\hat{c} = \left(\frac{c}{\sigma}\right)^2
\label{eq:tol norm}
\end{align}

and lambda is the sum of the squares of the position estimate again normalized by the standard deviation:

\begin{align}
\lambda(\boldsymbol{x}) = \sum_{j=1}^3 \left(\frac{x_j}{\sigma}\right)^2
\label{eq:lambda}
\end{align}

\begin{figure}[htbp]
\centering
\framebox{\parbox{3.4in}{
\includegraphics[width=0.48\textwidth]{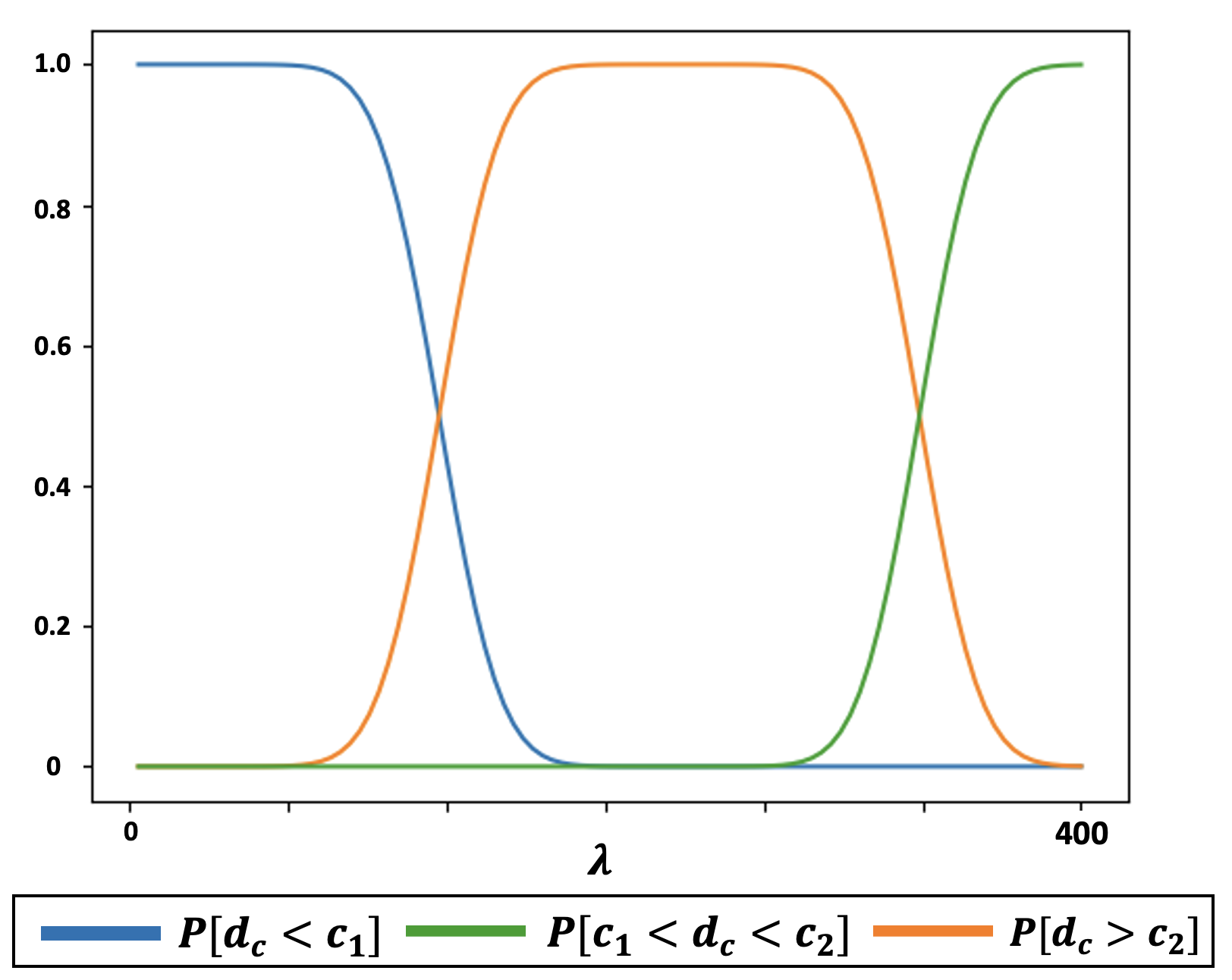}
        }}
\caption{The CDF function for each of the three defined inequalities.}
\label{fig:cdf}
\end{figure}

A plot of the CDF function for each of the three inequalities represented by $b$ can be seen in Figure \ref{fig:cdf}. The noncentral chi-squared distribution models the distribution of the sum of square of k independent, standard normally distributed variables. This is equal to the square of the euclidean distance of a noisy position estimate assuming that there is uniform standard deviation ($\sigma$) of variables across all axis. During testing, $\sigma$ can be computed by taking the standard deviation of the distance between the cat's position estimate and mouse's location over the last $N$ estimates.



\subsection{MPC control}

The RL policy provides a positional command, however, the necessary thrust control must still be computed in order to actuate the satellite in simulation and execute said command. An MPC was designed for this purpose. The controller attempts to minimize a cost function $J$ which is a function of the consumed fuel and deviation to goal given some control $\bold{u}$: 

\begin{align}
\min\limits_{\bold{u}_{0|k} \dots \bold{u}_{M-1|k}}  & J = \sum\limits_{i=0}^{M-1} (\bold{e}_{i|k}^TQ\bold{e}_{i|k} + \bold{u}_{i|k}^TR\bold{u}_{i|k}) + \bold{e}_{M|k}^TQ\bold{e}_{M|k} 
\label{eq:mpc1}
\\
\textrm{s.t.} \quad & \dot{\bold{x}}_{i|k} = A\bold{x}_{i|k} + B\bold{u}_{i|k} \\
  & \bold{x}_{0|k} = \bold{x}_k     \\
  & \bold{u}_i \in \begin{bmatrix}\bold{u}_{lb} \bold{u}_{ub}\end{bmatrix}
\label{eq:mpc2}
\end{align}

In equations \ref{eq:mpc1}-\ref{eq:mpc2} $\bold{e}$ is the state error, $Q$ and $R$ are weight matrices, $A$ and $B$ are the constant system and control matrices, and $\bold{u}_{lb}$ and $\bold{u}_{up}$ are the lower and upper bounds of the thrust control. The system and control matrices are derived using Equations \ref{eq:cwhx}-\ref{eq:cwhz}. The notation $i|k$ implies i steps past k, where k is the current state of the system. During implementation $M$ is set to $8$, thus giving 8 thrust inputs for the mouse. However, only the first computed input correlating to timestep $i=0$ is used. Once this is done, the thrust is applied to the satellite in simulation, and the environment is propagated a single timestep.

%% file: txt/baselines.tex
We test our method against two deferent baselines chosen to represent the different methodologies currently seen in satellite collision avoidance or evasion. The first is a single burn delta-v maneuver designed to ensure the spacecraft misses a potential obstacle by a minimum distance. The second is an optimization based algorithm that attempts to solve for the optimal position at each timestep with respect to the same reward function used to train the RL policy. In addition, because these methods are not designed for extremely noisy data we find it beneficial to pass all observations through an Extended Kalman Filter (EKF) before being fed to either algorithm. Observation variables that have been filtered by the EKF are notated by the superscript $+$. A more detailed explanation of the EKF process can be found in Appendix \ref{appendix:ekf}.

\subsection{Delta-V Optimization}

The first baseline implemented is a Delta-V Optimization method which will be refereed to as DVO in further sections. This method, adapted from Slater Et al. \cite{doi:10.2514/1.16812} uses a closed for solution of the Clohesy Whiltshire equations to compute the minimum delta-v required to reach a minimum distance from the origin of the Hill frame. This approach is representative of current industry standards, which generally compute the necessary delta-v to alter a spacecrafts orbit enough that the probability of collision is near zero \cite{BONNAL2020637,Patnala2024}.  The algorithm aims to minimize the cost function:

\begin{align}
J = \mid\Delta\boldsymbol{V}\mid^2 - 
\lambda[\Delta\boldsymbol{V}^T\phi_{12}^TP\phi_{12}\Delta\boldsymbol{V}-D^2]
\label{eq:dv_opt}
\end{align}

Where $\Delta\boldsymbol{V}$ is the delta-v vector, $D$ is the minimum miss distance of the obstacle, $P$ is a symmetric projection matrix:

\begin{align}
P = I - \boldsymbol{e}\boldsymbol{e}^T
\end{align}

where $\boldsymbol{e}$ is a unit vector in the direction of the velocity in the Hill frame after correction. $\phi_{12}$ is the closed form solution of the Clohessy-Whiltshire equations for some fixed time $t$ assuming a starting position at the origin:

\begin{align}
\phi_{12} = 
\begin{bmatrix}
    \boldsymbol{c}_t & 2-2\boldsymbol{c}_t & 0 \\
    2\boldsymbol{c}_t - 2 & 4\boldsymbol{s}_t & 0 \\
    0 & 0 & \boldsymbol{s}_t
\end{bmatrix}
\label{eq:dvo_phi12}
\end{align}

Finally, $\lambda$ is a Lagrange multiplier that is computed by taking the largest eigenvalue of $\phi_{12}^TP\phi_{12}$, and the magnitude of the optimal delta-v can be computed by:

\begin{align}
\mid\Delta\boldsymbol{V}\mid = \sqrt{\frac{D^2}{\lambda_{max}}}
\end{align}

In order to compute the optimal delta-v, the unit vector $\boldsymbol{e}$ must already be chosen. In order to do this, we implement a similar strategy to the Greedy Recursive Search in order to find the optimal control. We start with an initial $\boldsymbol{e}$ equal to the vector opposite of the direction from the mouse to the average position of the last several cat estimations. We then compute the optimal delta-v for a range of vectors within an angle tolerance of the initial guess. The vector that provides minimal fuel is the one chosen.


\subsection{Greedy Recursive Search}

The DVO algorithm well characterizes the majority of satellite avoidance methods currently utilized today. However, it was not necessarily designed to optimize the cat and mouse problem we are exploring, and thus we provide a second optimization based method more tailored to the problem at hand known as Greedy Recursive Search (GRS). 
Our GRS algorithm solves for the optimal position command to provide to the satellite by iterating through an increasingly more granular state space of potential positions within a given tolerance of the averaged cat position estimate. The algorithm attempts to optimize the same reward function used to train the RL policy described in Equation \ref{eq:rew}, and the full algorithm can be seen in Algorithm \ref{alg:grs}. 

\begin{algorithm}
\caption{Greedy Recursive Search}\label{alg:grs}
\begin{algorithmic}[1]
\Function {GRS}
{$\phi_{min},\phi_{max},\theta_{min},\theta_{max},\boldsymbol{o}_i$}
\State $\boldsymbol{\phi} = [\phi_{min},...,\phi_{max}]$, $\boldsymbol{\theta} = [\theta_{min},...,\theta_{max}]$
\State $r_{min} \gets \infty$
\State $\boldsymbol{g}_{opt} \gets None$
\For {$\phi \in \boldsymbol{\phi}$}
\For {{$\theta \in \boldsymbol{\theta}$}}
\State $x \gets d_m\boldsymbol{c}_\phi\boldsymbol{s}_\theta$
\State $y \gets d_m\boldsymbol{c}_\phi\boldsymbol{c}_\theta$
\State $z \gets d_m\boldsymbol{s}_\phi$
\State $\boldsymbol{g}^H \gets \frac{\sum_{j=i-N}^i{\boldsymbol{z}^H_{a,j}}^+}{N} + \begin{bmatrix}x & y & z\end{bmatrix}$
\State $r \gets 1-c_1||\boldsymbol{g}^H||-c_2f_i(\boldsymbol{g}^H)$
\If{$r < r_{min}$}
    \State $r_{min} \gets r$
    \State $\boldsymbol{g}_{opt} \gets \boldsymbol{g}^H$
\EndIf
\EndFor
\EndFor
\If{$max(\phi_{max}-\phi_{min},\theta_{max}-\theta_{min}) < tol$}
\Return $\boldsymbol{g}_{opt}$
\Else 
\State $\begin{bmatrix}x & y & z\end{bmatrix} = \boldsymbol{g}_{opt} - \frac{\sum_{j=i-N}^i{\boldsymbol{x}^H_{a,j}}^+}{N}$
\State $\phi \gets atan2(z,\sqrt{x^2+y^2})$
\State $\theta\gets atan2(x,y)$
\Return $\text{GRS}(\phi \pm \frac{\phi_{max}-\phi_{min}}{a},\theta \pm \frac{\theta_{max}-\theta_{min}}{a},\boldsymbol{o}_i)$
\EndIf
\EndFunction
\end{algorithmic}
\end{algorithm}

In Algorithm \ref{alg:grs} $\phi_{min},\phi_{max},\theta_{min},$ and $\theta_{max}$ represent angle ranges in polar coordinates, they are generally set to $0$ and $2\pi$ when the algorithm is called. The algorithm takes the average location of the cat's most recent position estimates, and compares the potential reward for a range of points that are distance $d_m$ from the mean position. $d_m$ was chosen by analytically experimenting with values greater than $c_1$ in order to allow preemptive evasion. In order to solve for the optimal position command, the function is run recursively to analyze a smaller subset of potential points outside of the optimal point found from the previous iteration. Additionally, if the position of the cat is outside of $c_2$, the mouse will return to the origin like the action space defined for our constrained RL policy.

%% file: txt/experiments.tex
\begin{figure}[htbp]
\centering
\framebox{\parbox{3.4in}{
\includegraphics[width=0.475\textwidth]{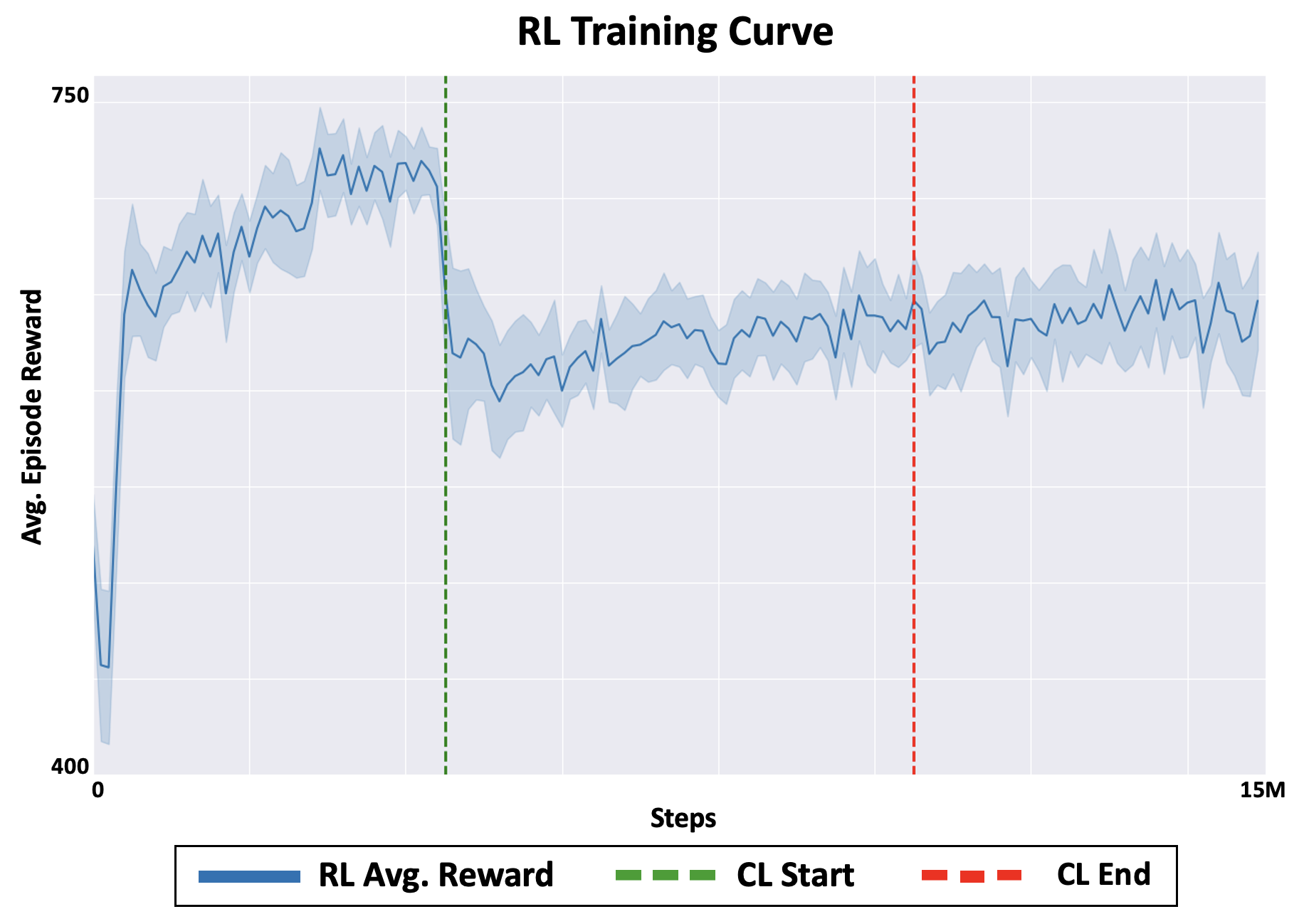}
        }}
\caption{A training curve of the RL policy with the start of the Curriculum Learning (CL) strategy marked in green, and the end marked in red.}
\label{fig:train}
\end{figure}

In this section we will discuss the implementation of our proposed method, real-world data used for experimentation, and the results of our experiments. We trained our policy using Soft Actor-Critic (SAC) which is an off-policy RL algorithm. We chose SAC for its entropy maximization strategy, which is well suited for learning in environments with large amounts of noise. During testing, 3 RL models were used -- each trained on the same environment with different seeds. A training curve of the RL model can be seen in Figure \ref{fig:train}. In both simulation and testing a constellation of 60 satellites was used, and the cat and mouse satellites each had a mass of 2,500kg. Mouse thrust was limited to 1N on each axis. During testing, cat behavior was determined using real-world data of unwarranted fly-bys, and RL performance was compared against a common satellite obstacle avoidance algorithm as well as a greedy optimization algorithm. 

\subsection{Testing Data}

In order to test the contained RL policy we use historical TLE data obtained from the United States Space Surveilance Network of two known fly-by events \cite{Space-Track.org}:

\begin{itemize}
    \item \textbf{Luch/Intelsat 37a October 29-31 2022}: Between mid October and mid Novermber of 2022, the Luch/Olymp satellite performed several orbital maneuvers to approach Intelsate 37e. At their closest they were only 8.2km appart. This was only one of numerous approaches Luch/Olymp performed on satellites in Geosynchronus orbit.
    \item \textbf{Luch2/Eutelsat Konnect VHTS January 14-16 2024 }: In mid January of 2024 the second Luch/Olymp satellite performed approaches toward several satellites on the Geosynchronus belt, the closest approach being only 4.5km with Eutelsat Konnect VHTS.
\end{itemize}

\begin{figure}[htbp]
\centering
\framebox{\parbox{3.4in}{
\includegraphics[width=0.475\textwidth]{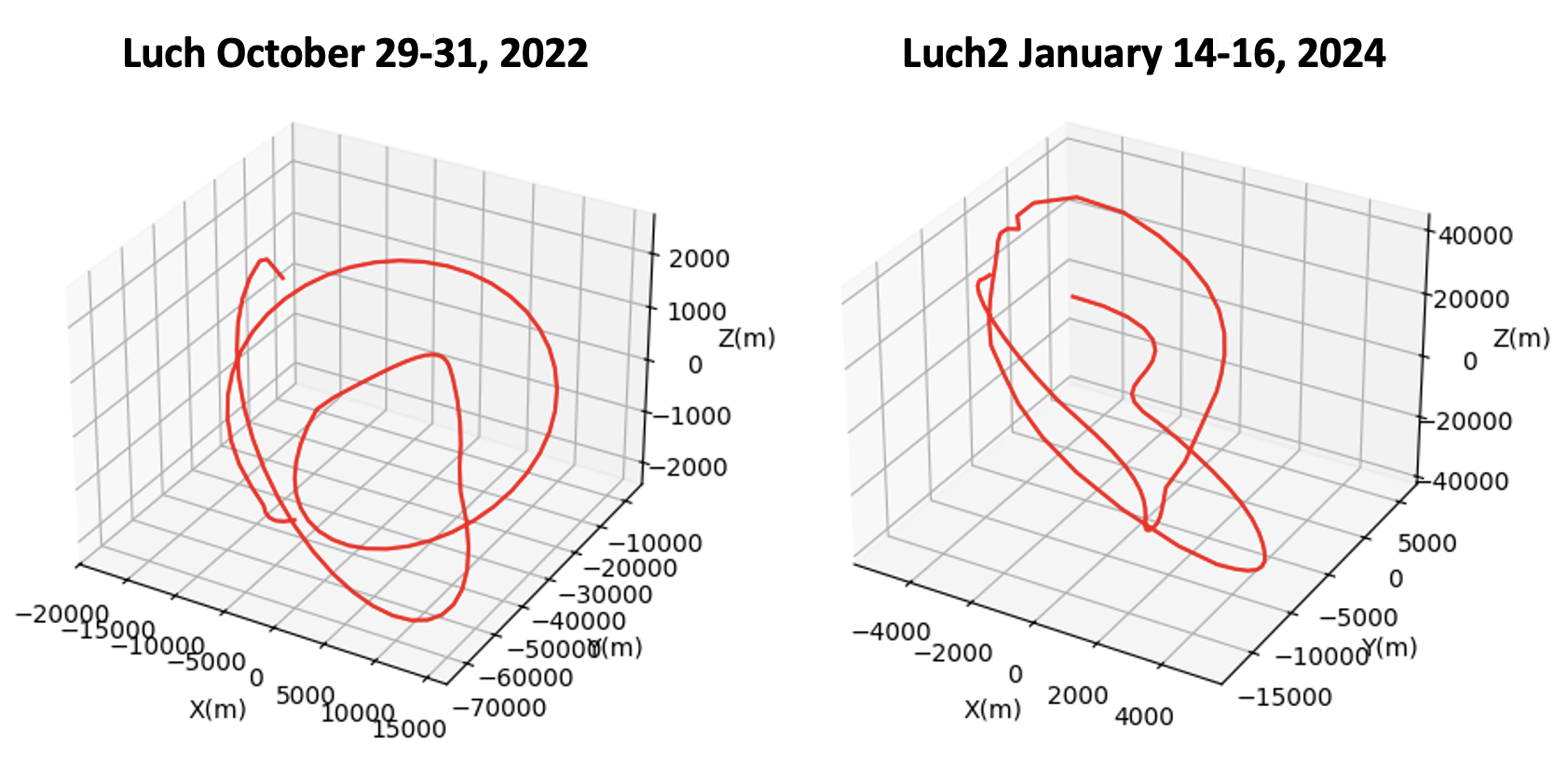}
        }}
\caption{The relative trajectories of Luch and Luch2 used during testing.}
\label{fig:luch trajs}
\end{figure}

For each event, the position of the non-cooperative space craft in relation to the vehicle it approached was computed and used as the cat's trajectory in simulation. The process of acquiring and processing the data is described in depth in Appendix \ref{alg:prop}. An image of each of these relative trajectories can be seen in Figure \ref{fig:luch trajs}.

Both our method's and the implemented baselines' performance during testing is evaluated using \ref{eq:rew}, which as discussed prioritizes distance from adversary, deviation from initial orbit, and fuel consumption. During testing, each method was given full control of the mouse vehicle, while the cat's trajectory was determined by the obtained TLE data. For each test scenario and method, results were averaged over 3 different seeds tested for 100 runs each.

%% file: txt/results.tex
\subsection{Results}

\begin{table}[!t]
\caption{Test Results\label{tab:res}}
\centering
\begin{tabular}{|c||c|c|}
\hline
Algorithm & Luch/Intelsat 37a & Luch2/Eutelsat\\
\hline
\hline
DVO & $135 \pm 30.6$ & $143 \pm 23.9$ \\
GRS & $807 \pm 23.3$ & $568 \pm 34.5$ \\
\textbf{RL} & $\boldsymbol{821 \pm 33.4}$ & $\boldsymbol{610 \pm 38.2}$ \\
\hline
\end{tabular}
\label{tab:res}
\end{table}

\begin{figure*}
    \framebox{\parbox{7in}{
    \includegraphics[width=\textwidth]{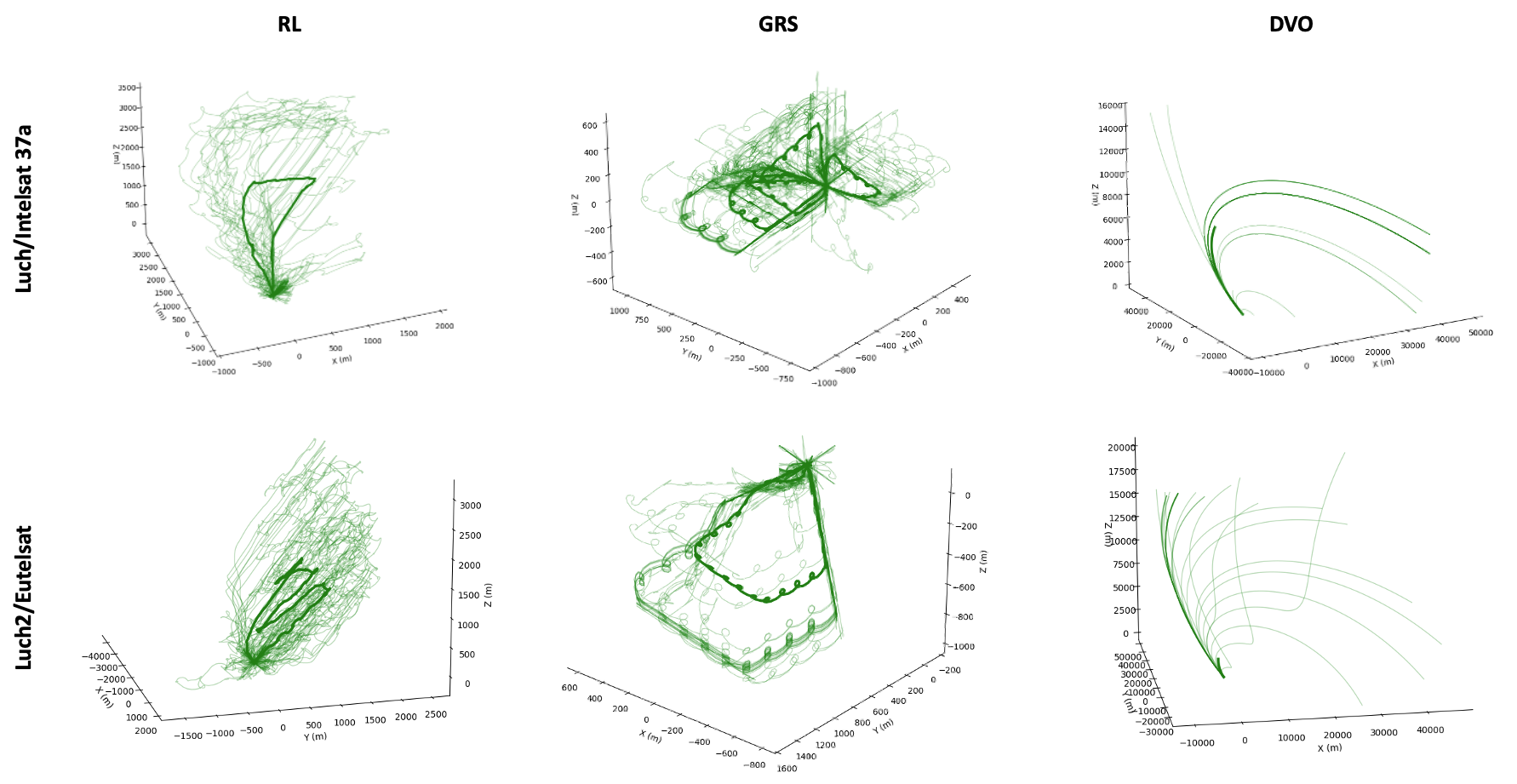}
            }}
  
  \caption{The trajectories generated by the RL policy, GRS, and DVO over 100 runs for both the Luch2/Eutelsat incident, and Luch/Intelsat37a incident. In the RL and GRS plots the average trajectory is marked by the bolded line.}
\label{fig:traj100}
\end{figure*}

Table \ref{tab:res} shows the averaged results and standard deviations of each algorithm for the two adversarial scenarios. It can be seen that the constrained RL policy clearly outperforms all other baselines. In theory, the GRS algorithm attempts to optimize the same reward function that the RL policy was trained on, and therefore should show near identical results. However, due to additional complexity from the dynamics of the environment and how this effects the sequential decision making process it under-performs the RL policy. Additionally, the DVO algorithm shows very low episode rewards, this is due to the fact that the episode ends after the mouse deviates from the initial orbit by more than 50km.

In addition to quantitative results, Figure \ref{fig:traj100} illustrates 100 different trajectories generated by each method for both adversarial scenarios. It is worth noting that in the image the RL policy's trajectory is subject to greater variation which is partially due to the fact that SAC samples its actions from a probability distribution and thus will not chose the same action when provided identical observations. Despite this greater variability in path taken it can be seen that the RL policy provides much smoother trajectories than the GRS algorithm, and has less variation in overall performance (Table \ref{tab:res}).

\begin{figure}
    \framebox{\parbox{3.4in}{
    \includegraphics[width=\columnwidth]{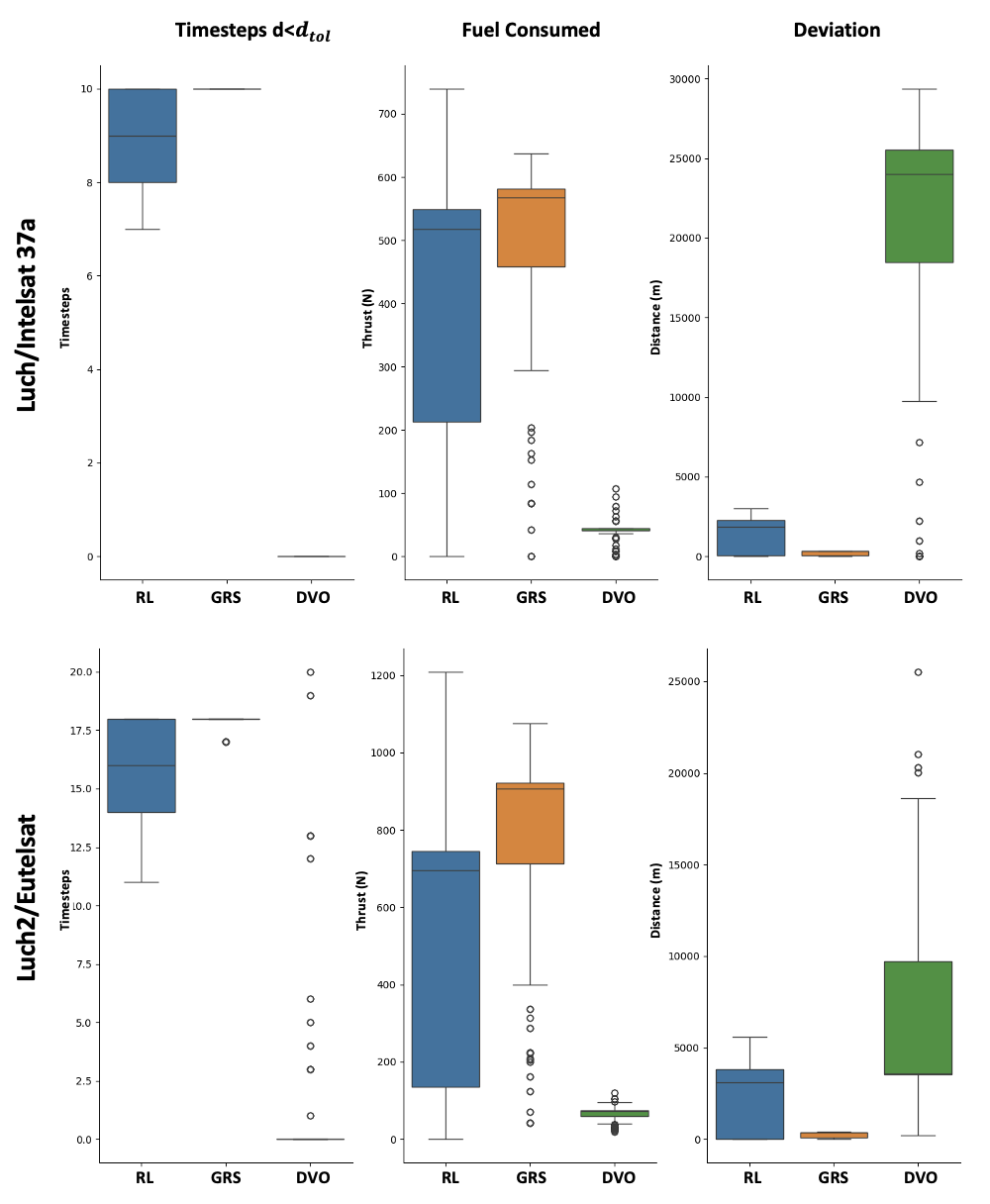}
            }}
  
  \caption{Box plots comparing the number of timesteps spent within $d_{tol}$ of the cat (left), the total fuel consumed in each episode (center), and average deviation from the initial orbit (right) for each control policy during testing. Results for both the Luch/Intelsat 37a event (top) and Luch/Eutelsat event (bottom) are provided.}
\label{fig:eval}
\end{figure}

In addition to the trajectories seen in Figure \ref{fig:tdoa error} we provide a thorough analysis into the performance of each control policy in regards to the same three parameters utilized in the reward scheme: timesteps spent within $d_{tol}$ of the cat, fuel consumed, and deviation from the origin. These results can be seen in Figure \ref{fig:eval}. Unsurprisingly, the DVO algorithm spends the least amount of timesteps within $d_{tol}$ of the cat, and consumes the least amount of fuel, however, its average deviation is extremely large. It should be noted that the average deviation of the DVO algorithm shown is in fact lower than the true average deviation because of episode cut-offs, thus, if each test was run for the full duration of the simulated cat trajectory the deviation would be significantly higher. Of more interest to this work is the comparison in performance between the RL policy and GRS algorithm which shows that the RL policy's success can be attributed to spending less time within $d_{tol}$ of the cat and consuming less fuel despite having a higher average deviation. 





%% file: txt/discussion.tex
The results in the previous section provide considerable insight into the capabilities and advantages of implementing different control schemes for satellite evasion. Most evident are the limitations of the DVO algorithm, which optimizes fuel and maintains a minimum distance from the uncooperative spacecraft, however, has no mechanism for maintaining minimum deviation from the initial orbit. This is exacerbated by the orbital dynamics: any deviation in the x or y axis of the local frame results in a change in orbit radius which will result in the spacecraft further deviating from its initial orbit with increasing acceleration unless there is some form of intervention. DVO is representative of many of today's standards for space-based avoidance, which were largely designed for systems that both lack maneuverability and sensing capabilities. 

\begin{table}[!t]
\caption{Localization Noise\label{tab:noise}}
\centering
\begin{tabular}{|c||c|c|c|}
\hline
Constellation Size & $\sigma_X$ (km) & $\sigma_Y$ (km) & $\sigma_Z$ (km)\\
\hline
\hline
30 & 5,542 & 5,173 & 2,045 \\
60 & 4.703 & 2.446 & 37.90 \\
100 & 0.239 & 0.003 & 0.147 \\
150 & 0.193 & 0.003 & 0.117 \\
200 & 0.165 & 0.004 & 0.099 \\
\hline
\end{tabular}
\label{tab:noise}
\end{table}

In addition to demonstrating limitations of current standards in satellite avoidance, we will highlight some of the key benefits to implementing a constrained Reinforcement Learning policy for satellite control compared to a more common greedy optimization algorithm. Primarily, how its formulation of the environment as an MDP allows it to better optimize control over a longer horizon. Perhaps this is best seen in Figure \ref{fig:traj100} when comparing the much smoother trajectories of the RL policy to that of the GRS algorithm. We attribute this to the fact that the GRS algorithm does not take into consideration the potential for maximizing performance given its future state after taking an action. Consequentially, the algorithm picks actions that do not allow for smooth maneuvering to future positional commands due to the dynamics of the simulation and capabilities of the MPC. This not only results in greater fuel cost, but also slows down the mouse's ability to travel larger distances in shorter periods of time: providing some insight to its low average deviation from the origin and high number of timesteps spent within $d_{tol}$ of the adversary. Additionally notable, while the RL policy outperformed the GRS algorithm in overall reward, timesteps spent in the vicinity of the cat, and fuel consumed, it did not outperform the GRS algorithm in average deviation from the origin. This may be due to several aspects of the GRS algorithm, including simply choosing optimal control that is not far from the origin, inadequate handling of noisy observation data, and slowed maneuvering due to shorter horizon planning as previously mentioned. 

While we have provided several key benefits and limitations to the different control schemes implemented, we also recognize that the performance of all methods can be improved by reducing the noise of the location estimates provided in the observation. Although the CRLB Bounds provided by Ho et al. give a strict limit on the performance of any TDOA localization algorithm, the accuracy of the algorithm can be improved by increasing the number of RF sensors. For this experiment, we chose a constellation of 60 satellites which results in approximately 10 to 25 LEO satellites within a $2^\circ$ beam transmitted from GEO at any given time. The size of the constellation was selected to reflect a feasible number of assets given current constellation sizes as well as a manageable noise level. In table \ref{tab:noise} we have provided the average CRLB bounds of a number of different constellation sizes in order to compare the localization capabilities of the 60-asset constellation to others. Once a constellation approaches 100 assets the localization accuracy can be brought down to within 100m, however, this is unrealistic given current satellite constellation magnitudes. A constellation that is too small provides error, which is insufficient for any type of evasion. Finally, a constellation of 60 provides a sufficient tradeoff. While more finite accuracies are more desirable, we have shown it is sufficient for intelligent evasion. Additionally, it should be noted that the large Error in the Z direction is generally due to outliers which are multiple magnitudes greater than the average noise as shown in Figure \ref{fig:tdoa error}. While uncommon these outliers occur due to the configuration of sensors and Dilution of Precision (DOP).




%% file: txt/conclusion.tex
Our work proposes an architecture for improved SDA and evasive control by leveraging RF informed localization techniques and a constrained RL control scheme. In this work we have provided the necessary simulation design for training such control schemes, and validated our work by testing on real-world, open data about uncooperative satellite maneuvers. Finally, we compare the constrained RL policy's performance to an optimization-based algorithm tailored for cat and mouse games, as well as a standard obstacle avoidance algorithm common in satellite avoidance today. We demonstrated that neither alternative outperforms the RL policy on established performance metrics. Our work highlights several key advantages to deploying RL methods in the space environment. First we establish RL's ability to use noisy, but ubiquitous RF data to establish a robust control policy and second, we establish RL's ability to learn and navigate the complex dynamics of the space environment. Furthermore, RL provides the additional benefit of being computationally light weight: the model used during experimentation is comprised of 2 layers of 256 nodes, and takes less time to run on a CPU than either optimization algorithm implemented. As the space domain continues to develop and become more congested the current SDA capabilities and planning options will not provide the necessary reactivity and mobility required of future spacecraft. Space-based RF sensing unlocks a new frontier in satellite autonomy capabilities; when integrated with advanced control algorithms — such as RL-based policies — it has the potential to expand the boundaries and improve the accessibility of autonomous satellite maneuvering.

%% file: txt/appendix.tex
\begin{appendices}
\section{Extended Kalman Filter}
\label{appendix:ekf}







As mentioned in Section \ref{sec:baselines}, the position estimates provided by the TDOA localization algorithm are extremely noisy. In order to mitigate errors in our baselines caused by this an EKF is implemented in order to smooth the position data before it is fed as intput to either of the baseline control algorithms. The EKF implemented follows an architecture similar to that often seen in space systems \cite{springerSatelliteOrbits}. At each timestep, the EKF must start with a reference state vector from the previous timestep $\boldsymbol{x}(t_{i-t})$, and covariance matrix $\boldsymbol{P}_{i-1}$. For the following equations governing the EKF, the state vector $\boldsymbol{x}$ is defined as:

\begin{align}
\boldsymbol{x} = \begin{bmatrix}
    x^E \\
    y^E \\ 
    z^E \\
    v_x^E \\
    v_y^E \\
    v_z^E
\end{bmatrix}  
\end{align}

We then predict the state vector and covariance matrix of the current timestep using the following equations:

\begin{align}
\boldsymbol{x}_i^- = \boldsymbol{x}(\boldsymbol{x}(t_{i-t}))
\end{align}

\begin{align}
\boldsymbol{P}_{i}^- = \boldsymbol{\Phi}_i\boldsymbol{P}_{i-1}\boldsymbol{\Phi}_i^T
\end{align}

Where the function $\boldsymbol{x()}$ represents Cowell's Formulation, which propagates the state forward a single timestep using Newton's Equations of motion:

\begin{align}
\boldsymbol{F} = \frac{Gm_1m_2}{r^3}\hat{\boldsymbol{r}}
\label{eq:grav}
\end{align}

And, $\boldsymbol{\Phi}$ is the state transition matrix computed from a second order Taylor expansion series:

\begin{align}
\boldsymbol{\Phi} = \boldsymbol{I} + \boldsymbol{F}\delta t + \boldsymbol{F}^2\frac{\delta t}{2}
\end{align}

where $\boldsymbol{F}$ is the state matrix of the 2-body problem $\frac{\partial \dot{\boldsymbol{X}}}{\partial \boldsymbol{X}}$ derived from equation \ref{eq:grav}. The Kalman gain $\boldsymbol{K}$ is then updated:

\begin{align}
\boldsymbol{K}_i = \boldsymbol{P}_i^-\boldsymbol{H}_i^T(\boldsymbol{H}_i\boldsymbol{P}_i^-\boldsymbol{H}_i^T + \boldsymbol{R}_i^{-1})^{-1}
\end{align}

where $\boldsymbol{H}$ is the measurement model which during experimentation and testing is equal to the identity matrix. Once the Kalman gain is computed, we can update the predicted state vector:

\begin{align}
{\boldsymbol{x}^E_{a,i}}^+ = {\boldsymbol{x}^E_{a,i}}^- + \boldsymbol{K}_i
({\boldsymbol{x}^E_{a,i}}^\prime - \boldsymbol{H}_i{\boldsymbol{x}^E_{a,i}}^-)
\end{align}

This equation is nearly identical to Equation \ref{eq:ekf1} in the Methods section, however, we show its full form here where $\boldsymbol{H}_i$ is not taken out. Finally, the covariance matrix is then updated in preparation for the next update:

\begin{align}
\boldsymbol{P}_i^+ = (1 - \boldsymbol{K}_i\boldsymbol{H}_i)\boldsymbol{P}_i^-
\end{align}

The updated state vector ${\boldsymbol{x}^E_{a,i}}^+$ and covariance matrix $\boldsymbol{P}_i^+$ are then used as the state vector and covariance matrix from the previous timestep in the next update. The position of the updated state vector ${\boldsymbol{x}^E_{a,i}}^+$ is then translated into the Hill frame following Equation \ref{eq:translate} in order to provide ${\boldsymbol{x}^H_{a,i}}^+$ to the control algorithm.

Additionally, during initialization $\boldsymbol{P}$ is a diagonal matrix of the square of the average error standard deviations of the constellation's position estimates found in Table \ref{tab:noise}, and $R$ was tuned through trial and error to provide the most accurate results.

\section{Real-World Data Acquisition}

Both data sets used for the Luch/Intelsate37a and Luch2/Eutelsat Konnect incidents were obtained from the U.S. Space Surveillance Network (SSN), and required minor processing in order to be used for testing. In this section we will discuss all steps involved. The initial data can be accessed through Space-Track.org \cite{Space-Track.org}. The data is received in the form of Two-Line Elements (TLE), which provides an update of a satellites ephemeris data one to several times per day. In order to compute a full trajectory of the spacecraft the orbit must be propagated from each TLE data point for the duration of time before the next recorded TLE data point. Each propagation can be defined as a set of points in the ECEF frame at constant timestep $\delta t = 3s$ appart:

\begin{align}
\boldsymbol{p}_{s,j,k} = [\boldsymbol{x}_{s,0}^E,...,\boldsymbol{x}_{s,k}^E]
\label{eq:propagation}
\end{align}

Where $\boldsymbol{x}_{x,i}^E$ is a point in the trajectory at timestep i which ranges between 0 and $k$. The subscripts of $\boldsymbol{p}_{s,j,k}$ represent the spacecraft being propagated, the index of the TLE data point used to initialize the propagation, and the number of timesteps in the propagation respectively. It should be noted that $k$ is unique for each propagation as the time in between TLE data points is not consistent These propagations can then be concatenated together to provide a full trajectory $\boldsymbol{P}_s$. In order to compute each propagation we use the SGP4 propogator developed by the US Space Force (USSF) also released through Space-Track.org.

\begin{algorithm}
\caption{Propagation Discontinuity Adjustment}\label{alg:prop}
\begin{algorithmic}[1]
\State Given $\boldsymbol{P}_s$
\For{$\boldsymbol{p}_{s,j,k}$ in $\boldsymbol{P}_s$}
\If{$j \neq len(\boldsymbol{P}_s)$}
\State $e = \boldsymbol{p}_{s,j+1,k}[0] - \boldsymbol{p}_{s,j,k}[k]$
\For{$\boldsymbol{x}_{x,i}^E$ in $\boldsymbol{p}_{s,j,k}$}
\State $\boldsymbol{x}_{x,i}^E$ = $\boldsymbol{x}_{x,i}^E + e \frac{i*\delta t}{k*\delta t}$
\EndFor
\EndIf
\EndFor
\end{algorithmic}
\end{algorithm}

The trajectory provided using this method will have discontinuities between the stitched together propagations primarily due to satellite maneuvers that may have occurred between TLE readings which is not taken into account when propagating the orbit, but also orbital perturbations. In order to fix this the propagation is smoothed by adding an error term to each value of the propagation, this term is computed by linearly interpolating between 0 and the discontinuity between the propagation final point in cartesian coordinates the the proceeding propagation initial point as described in Algorithm \ref{alg:prop}. This method assumes that the maneuver causing the discontinuity was performed as a single impulsive maneuver.

Finally, once a smooth trajectory of both spacecraft is computed the trajectory of the unwarranted or cat spacecraft must be recorded in the local Hill frame relative to the mouse. The cat's trajectory is first interpolated over time using a spline interpolation function in order to compute the cat's trajectory in the same timestamps as the mouse's. Then, for each timestep the orbital parameters of the mouse spacecraft is computed from Kepler's equations given its position and computed velocity. Due to the fact that the timestep is in an order of seconds, the velocity at each point is computed by assuming linear motion:

\begin{align}
\boldsymbol{v}_{x,i}^E = \frac{\boldsymbol{x}_{x,i+1}^E-\boldsymbol{x}_{x,i}^E}{\delta t}
\label{eq:data_vel}
\end{align}

Additionally, since all spacecraft trajectories used are in GEO, their orbits are assumed to be cirular. Once the orbital parameters are computed, we know the local Hill frame of the mouse for this specific timestep. The cat's position in this frame is then computed using Equation \ref{eq:translate} for the corresponding point in the cat's trajectory. Once this is done for each timestep in the trajectory, we have the real-world trajectory of the unwarrented spacecraft in relation to the mouse over a fixed period of time.

\end{appendices}